\documentclass[runningheads]{llncs}

\usepackage{eccv}

\usepackage{eccvabbrv}

\usepackage{graphicx}
\usepackage{booktabs}
\usepackage{multirow}
\usepackage{adjustbox}
\usepackage{pifont}
\usepackage{fontawesome} 
\usepackage{amsmath}
\usepackage{caption}
\usepackage{multicol}
\usepackage{color, xcolor}
\usepackage{hyperref}
\usepackage{color, colortbl}
\hypersetup{pdfstartview=FitH,
            colorlinks=true,
            linkcolor=red,
            anchorcolor=blue,
            }
\definecolor{mygray}{gray}{.88}
\usepackage[accsupp]{axessibility}  

\usepackage{hyperref}

\usepackage{orcidlink}
\usepackage{marvosym}

\begin{document}

\title{Spiking Wavelet Transformer} 

\titlerunning{Spiking Wavelet Transformer}

\author{Yuetong Fang\inst{1}$^\dagger$\orcidlink{0000-0003-0228-9082} \and
Ziqing Wang\inst{1, 2}$^\dagger$\orcidlink{0009-0004-8940-0461} \and
Lingfeng Zhang\inst{1}\orcidlink{0009-0006-0696-4363} \and Jiahang Cao\inst{1} \and Honglei Chen\inst{1} \and Renjing Xu\inst{1}$^{\href{mailto:renjingxu@hkust-gz.edu.cn}{\textrm{\Letter}}}$\orcidlink{0000-0002-0792-8974}
\thanks{$^\dagger$~Equal Contribution; $^\textrm{\Letter}$~ Corresponding Author. }}
\authorrunning{Y. Fang, Z. Wang et al.}
\institute{The Hong Kong University of Science and Technology (Guangzhou), China\\
Northwestern University, IL, USA\\
\email{renjingxu@hkust-gz.edu.cn}\\
}


\maketitle


\begin{abstract}
   Spiking neural networks (SNNs) offer an energy-efficient alternative to conventional deep learning by emulating the event-driven processing manner of the brain. Incorporating Transformers with SNNs has shown promise for accuracy.
   However, they struggle to learn high-frequency patterns, such as moving edges and pixel-level brightness changes, because they rely on the global self-attention mechanism. Learning these high-frequency representations is challenging but essential for SNN-based event-driven vision. To address this issue, we propose the Spiking Wavelet Transformer (SWformer), an attention-free architecture that effectively learns comprehensive spatial-frequency features in a spike-driven manner by leveraging the sparse wavelet transform.
    The critical component is a Frequency-Aware Token Mixer (FATM) with three branches: 1) spiking wavelet learner for spatial-frequency domain learning, 2) convolution-based learner for spatial feature extraction, and 3) spiking pointwise convolution for cross-channel information aggregation - with negative spike dynamics incorporated in 1) to enhance frequency representation. The FATM enables the SWformer to outperform vanilla Spiking Transformers in capturing high-frequency visual components, as evidenced by our empirical results. Experiments on both static and neuromorphic datasets demonstrate SWformer's effectiveness in capturing spatial-frequency patterns in a multiplication-free and event-driven fashion, outperforming state-of-the-art SNNs. SWformer achieves 
    a 22.03\% reduction in parameter count, and a 2.52\% performance improvement on the ImageNet dataset compared to vanilla Spiking Transformers. The code is available at: \url{https://github.com/bic-L/Spiking-Wavelet-Transformer}.

  \keywords{Spiking Neural Networks \and Wavelet Transform \and Vision Transformer \and Event-based vision}
\end{abstract}

\section{Introduction}
\label{sec:intro}

Spiking neural networks (SNNs) have gained considerable interest as a promising alternative to standard artificial neural networks (ANNs)~\cite{he2023weaklysupervised, ye2022perceiving, wu2023mask, he2024diffusion}. Inspired by biological neurons, SNNs process information via binary events called spikes. Neurons transmit spikes only when their accumulated membrane potential exceeds a firing threshold, otherwise remaining inactive~\cite{roy2019towards}. This sparse, event-driven processing offers orders of magnitude gains in efficiency and performance over conventional computing paradigms, especially on low-power neuromorphic chips, such as Loihi~\cite{davies2018loihi}, True North~\cite{merollaMillionSpikingneuronIntegrated2014}, and Tianjic~\cite{pei2019towards}, which compute spikes asynchronously. With these advantages, there is a growing body of research applying SNNs, such as classification~\cite{mengTrainingHighPerformanceLowLatency2022, wang2023masked, zhou2022spikformer}, object detection~\cite{caoSpikingDeepConvolutional2015, su2023deep}, autonomous driving~\cite{zhu2024autonomous} and tracking~\cite{yangDashNetHybridArtificial2019, ji2023sctn}. Despite the energy efficiency of SNNs, they lag behind ANNs in terms of accuracy, posing a major challenge.

\begin{figure*}[t]
 
 \hspace{0.5cm}\includegraphics[width=0.9\linewidth]{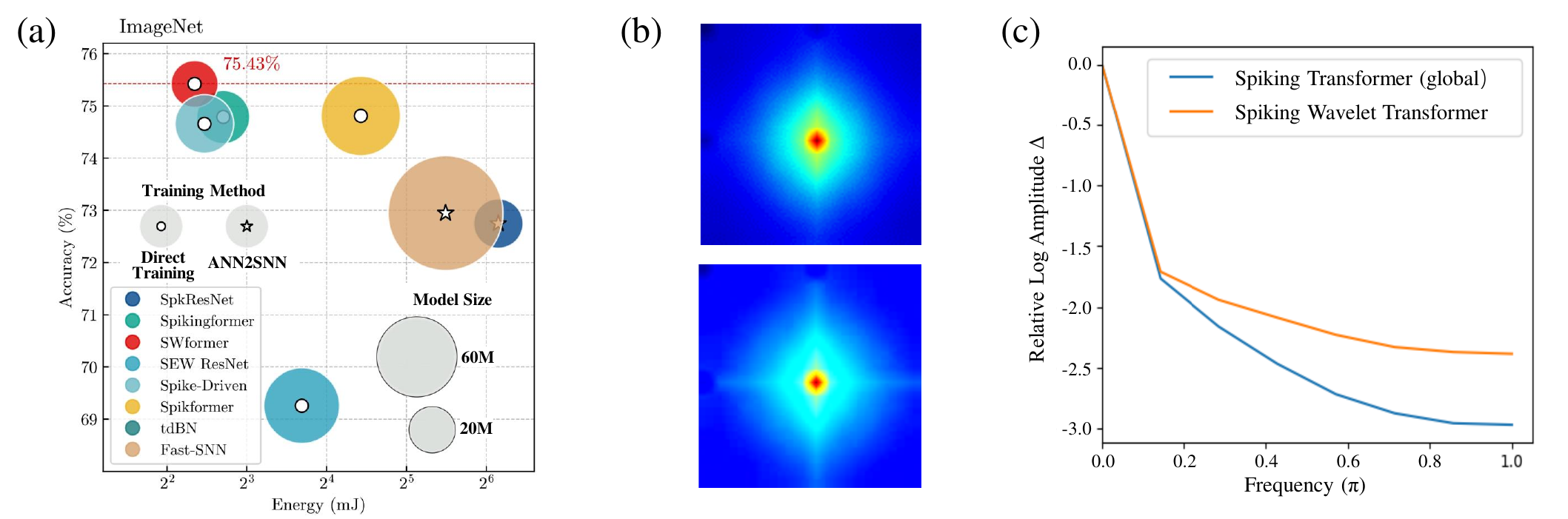}
 \caption{{
  (a) Performance of SWformer and other SOTA SNN models in top-1 accuracy and energy consumption (detail in supplementary), with marker size reflecting model size.
 (b) Fourier spectrum comparison between the Spiking Transformer with global attention~\cite{yao2023spike} (top) and SWformer (bottom). Brighter colors indicate higher magnitudes. (c) Corresponding relative log amplitudes of Fourier-transformed feature maps. {(b-c) show SWformer captures more high-frequency signals, leading to better performance.}}}
 \label{fig:main}
\end{figure*}

To get the best of both worlds, a line of works focuses on incorporating advanced architectures from ANNs with the unique spiking mechanism in SNNs. This has led to notable developments. The introduction of residual learning into SNNs has facilitated the development of deeper network architectures and thus enhanced their performance~\cite{fangDeepResidualLearning2021, zhengGoingDeeperDirectlytrained2021, hu2021spiking}. More recently, integrating attention mechanisms has granted SNNs improved global information capturing, strengthening their capability to handle intricate patterns~\cite{yao2021temporal, zhangSpikingTransformersEventBased2022a}. This success has motivated researchers to discover the potential of combining powerful Transformer architecture with energy-efficient SNNs~\cite{zhou2022spikformer, yao2023spike}. While there has been some research in this direction, existing works mostly inherited the architecture from Vision Transformer~\cite{vaswaniAttentionAllYou2017}, known to function solely in the spatial domain and exhibit similar characteristics of low-pass filters~\cite{park2022vision, chen2023sparse}. 

As a representative branch in neuromorphic computing, SNNs mimic biological vision by continuously sampling the input data and independently generating spikes in response to changes in the visual scene~\cite{hopkins2018spiking}, conveying abundant local information. Specifically, neuromorphic data captures only brightness changes, primarily moving edges, which represent high-frequency patterns~\cite{li2022neuromorphic}. As supported by the empirical comparisons in Fig.~\ref{fig:main}(b-c), though Spiking Transformers are highly capable of handling low-frequency components, like global shapes and structures, they are not very powerful for learning high-frequency information, mainly including abrupt changes in images such as local edges and textures~\cite{si2022inception, chen2023msp} - this is intuitive since self-attention, their primary mechanism is a global operation that aggregates information across non-overlapping image patches. Porting the frequency information into SNNs is a natural and appealing idea; however, this has been non-trivial due to the spike-driven nature of SNNs. 

Frequency analysis methods, like the Fourier transform, rely on precise matrix multiplications~\cite{bochner1949fourier}, while SNNs use sparse, binary signaling with only a portion of neurons activated at any given time~\cite{davies2018loihi, wang2023bursting}. This sparse, binary signaling mechanism presents a significant obstacle in devising a spiking equivalent to measure the frequency features accurately. To reduce the information loss, existing works have investigated the adoption of precise data encoding like time-to-value mapping~\cite{lopez2021spiking, lopez2022time}, though at the expense of high latency. We argue that time-frequency decomposition can be a more effective and efficient representation space for SNNs, considering their sparse and robust properties~\cite{li2020wavelet, he2023camouflaged}. In fact, the human visual system discerns elementary features through time-frequency components~\cite{Gaudart:93, lee1994wavelet}: it is found that the human visual system analyzes images in a way similar to the multi-resolution breakdown by the wavelet functions.

We propose the Spiking Wavelet Transformer (SWformer) to effectively capture time-frequency information in an event-driven manner. SWformer integrates the robustness of wavelet transforms with the energy efficiency of Spiking Transformers. As shown in Fig.~\ref{fig:main}, SWformer captures more high-frequency information than Spiking Transformers with global attention, significantly enhancing performance. It processes data in a multiplication-free, event-driven way compatible with neuromorphic hardware while effectively capturing spatial-frequency information. The main contributions of this paper are:

\begin{itemize}
    
    \item We propose SWformer, a novel attention-free architecture that integrates time-frequency information with Spiking Transformers, enabling feature perception across a wide frequency range in an event-driven manner.
    
    \item A key component of SWformer is the Frequency-Aware Token Mixer (FATM), which processes input in three branches to learn spatial, frequency, and cross-channel representations, allowing it to capture more high-frequency visual information than vanilla Spiking Transformers.
    
    \item We incorporate negative spike dynamics, a simple yet effective method supported by theoretical and experimental observations, to provide robust frequency representation in SNNs.

    \item Extensive experiments show that our model significantly outperforms SOTA SNN performances, achieving a 2.95\% improvement on static datasets like ImageNet and a remarkable 4\% increase on neuromorphic datasets, such as CIFAR10-DVS.
    
\end{itemize}

\section{Preliminary}

\subsection{Bio-inspired Spiking Neural Networks}
SNNs are variants of ANNs that mimic the spatial-temporal dynamics and binary spike activations found in biological neurons~\cite{roy2019towards, yao2023spike}. This spike-based temporal processing paradigm allows sparse while efficient information transfer. 
However, the non-differentiable spike function hinders the use of gradient-based backpropagation to train SNNs effectively. 
Two main solutions exist: ANN-to-SNN conversion~\cite{buOptimalANNSNNConversion2021, dingOptimalAnnsnnConversion2021} and direct training~\cite{zhengGoingDeeperDirectlytrained2021, meng2022training}. The ANN-to-SNN conversion method aims to bridge the continuous activation value of ANNs with the firing rate of SNNs through neuron equivalence~\cite{buOptimalANNSNNConversion2021, dingOptimalAnnsnnConversion2021}, borrowing backpropagation to achieve high performance but requiring long simulation timesteps and high energy consumption.
In this work, we employ the direct training method to fully leverage the benefits of low-power and sparse event-driven computing of SNNs.

\subsection{Neuromorphic Chips}
Neuromorphic chips, inspired by the brain, merge processing and memory units, using spiking neurons and synapses as fundamental elements~\cite{roy2019towards, schuman2022opportunities}. As shown in Fig.~\ref{fig:neuro}, the synapse block processes incoming spikes, retrieves synaptic weights from memory and generates spike messages to be routed to other cores~\cite{merollaMillionSpikingneuronIntegrated2014, davies2018loihi, pei2019towards}. This replaces energy-consuming Conv and MLP operations with energy-efficient routing and sparse addition algorithms~\cite{davies2021advancing, ye2023lacera}, though self-attention is not yet supported. The spike-based computation grants neuromorphic chips high parallelism, scalability, and low power consumption (tens to hundreds of milliwatts)~\cite{basu2022spiking}. Our SWformer design adheres to the spike-driven paradigm, making it well-suited for implementation on neuromorphic chips.

\hspace{1.8cm}\includegraphics[width=0.6\linewidth]{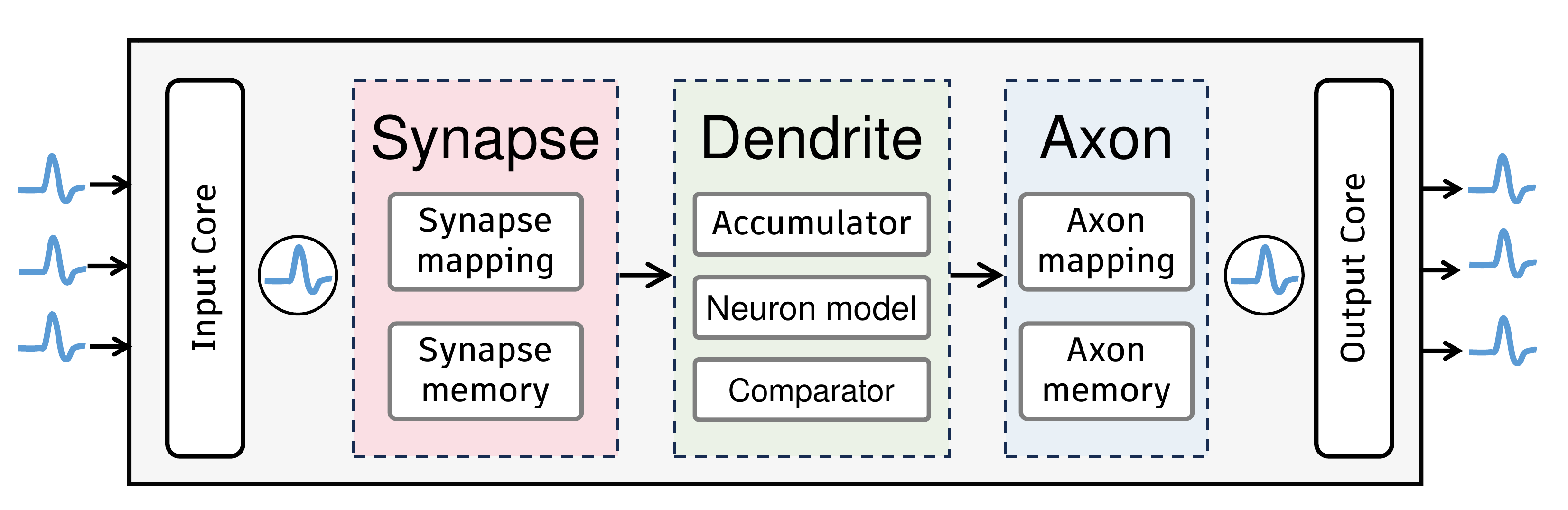}
\captionof{figure}{Processing flow of a synapse block. Neuromorphic chips follow a spike-based computation paradigm, where both inputs and outputs are in spike form.~\cite{davies2018loihi}}
\label{fig:neuro}

\subsection{Spiking Vision Transformers}
Recent advancements in SNN architectures, inspired by deep learning~\cite{roy2019towards, yao2023attention, yang2024genuine} and brain-like processes such as long short-term memory and attention~\cite{rao2022long, bellec2018long}, have significantly improved their performance with the benefits of spike-driven processing.
 This progress has led to the creation of Spiking Transformers~\cite{zhou2022spikformer, yao2023spike, zhou2023spikingformer}, which merges the effectiveness of Transformers with the energy efficiency of SNNs, providing a solution for energy-sensitive scenarios~\cite{basu2022spiking}. 
However, previous works have directly inherited the Vision Transformer architecture~\cite{vaswaniAttentionAllYou2017}, whose core self-attention mechanism primarily focuses on low-frequency information through global exchange among non-overlapping patch tokens and neglecting high-frequency components like detailed information, local edges, and abrupt pixel-level changes~\cite{si2022inception}. 
Our work emphasizes the importance of high frequencies for SNNs, which is expected given the independent and sparse spike generation of SNNs that yields abundant high-frequency data cross layers.

\subsection{Learning in the Frequency Domain}
Integrating frequency representation into SNNs is particularly important, considering neuromorphic data reflects brightness changes corresponding to high frequencies. Additionally, while static images do not inherently contain high-frequency information, spiking neurons can encode inputs into pixel-level brightness changes, enriching the images processed by the spiking layer with frequency information (Sec~\ref{sec:4.4}). 
However, few works have applied frequency representation to SNNs, such as devising spiking band-pass filters~\cite{jimenez2016binaural} or neurons that spike at specific frequencies~\cite{auge2020resonate}, struggling to capture full-frequency spectra.
More recently, Lopez et al.~\cite{lopez2021spiking, lopez2022time} adopted time-to-value mapping for accurate Fourier transform but at the cost of high latency~(\textasciitilde1024 timesteps). In this work, we ingeniously combine the sparsity of wavelet transform with the binary and sparse signaling of SNNs to provide robust frequency representation.



\section{Spiking Wavelet Transformer}

We devise the SWformer, a novel attention-free architecture that combines time-frequency information with Spiking Transformers. This allows for efficient feature perception across a wide frequency range without multiplication and in an event-driven manner. 
We first briefly introduce the spiking neuron layer, followed by an overview of SWformer and its components.

The spiking neuron layer encodes spatio-temporal information into membrane potentials, converts them into binary spikes, and passes them on to the next layer for continued spike-based computation. Throughout this work, we consistently use the Leaky Integrate-and-Fire (LIF) neuron model~\cite{maass1997networks}, as it efficiently simulates biological neuron dynamics. The following equations govern the dynamics of the LIF layer:

\begin{align}
U[n] & = V[n-1] + I[n], \\
s[n] & = H(U[n-1] - V_{\text{th}}), \\
V[n] & = V_{\text{reset}}s[n] + (\beta U[n])(1-s[n]),
\end{align}

At each timestep $n$, the current membrane potential $U[n]$ is generated by integrating the spatial input $I[n]$ from input data or intermediate operations like Conv and MLP, with temporal dynamics $V[n]$, which track the membrane potential over time. If $U[n]$ exceeds the threshold $V_{\text{th}}$, the neuron fires a spike ($s[n]$=1), otherwise it remains inactive ($s[n]$=0). The Heaviside step function $H(\cdot)$ determines spiking, where $H(x)$ = 1 when $x \geq 0$. The temporal output $V[n]$ updates based on the spiking activity and decay factor $\beta$. If the neuron does not fire, $U[n]$ decays to $V[n]$.


\hspace{0.0cm}\includegraphics[width=1.0\linewidth]{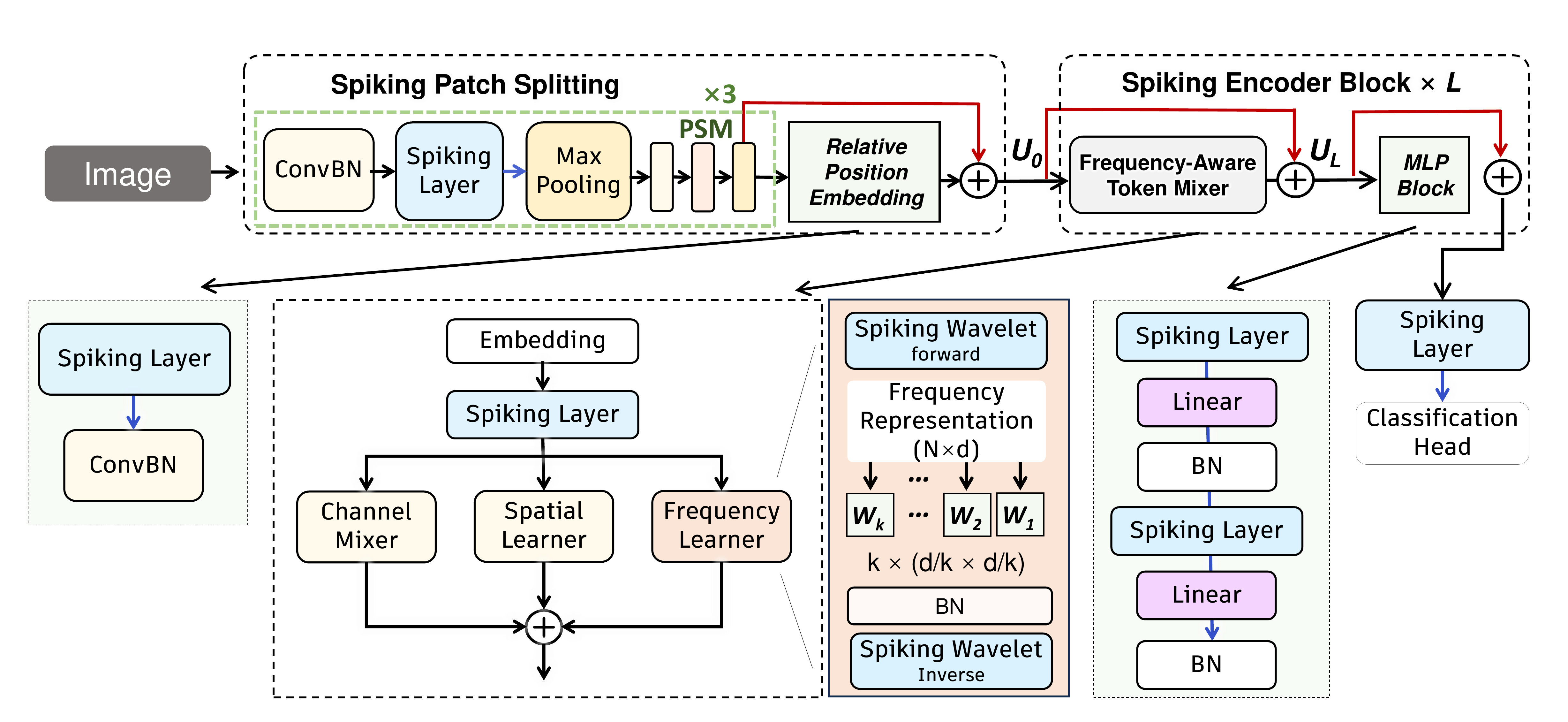}
\captionof{figure}{The overview of SWformer. 
 We present two main innovations inspired by~\cite{zhou2022spikformer}. Firstly, 
 FATM improves frequency perception in Spiking Transformers using only Conv and MLP operations, ensuring compatibility with neuromorphic hardware. Second, our Frequency Learner (FL) efficiently captures spectral features through spiking frequency representation and block-diagonal multiplication. ConvBN: a Conv layer followed by a BN layer.}
\label{fig:overview}

\subsection{Overall Architecture}

Fig.~\ref{fig:overview} presents SWformer, which comprises a Spiking Patch Splitting (SPS) module, Spiking Encoder blocks, and a linear classification head. The SPS module, based on the design in~\cite{zhou2022spikformer}, includes a Patch Splitting Module (PSM) with the initial four spiking Conv layers. For SNNs, the input sequence dimension is $I \in \mathbb{R}^{T \times C \times H \times W}$, where $T$ is the number of timesteps. In static datasets, images are repeated $T$ for creating a temporal sequence, while neuromorphic datasets inherently split data into $T$ frame sequences. For a 2D image sequence $I \in \mathbb{R}^{T \times C \times H \times W}$, the SPS is formulated as:
\begin{equation}
\begin{array}{ll}
U = \operatorname{PSM}(I), ~~& I \in \mathbb{R}^{T \times C \times H \times W}, U \in \mathbb{R}^{T \times N \times D} \\ [0.3em] 
s = \operatorname{\textit{Spk}}(U),  ~~& s \in \mathbb{R}^{T \times N \times D} \\ [0.3em] 
\operatorname{RPE} = \operatorname{ConvBN}(s), ~~& \operatorname{RPE} \in \mathbb{R}^{T \times N \times D} \\ [0.3em] 
U_0 = U + \operatorname{RPE}, ~~& U_0 \in \mathbb{R}^{T \times N \times D}
\end{array}
\end{equation}
\noindent where U and U$_0$ denote output membrane potential tensor of the PSM and SPS, respectively, with \textit{Spk}(·) symbolizing the spiking neuron layer. The resulting patches are processed by $l$ Spiking Encoder Blocks, each containing a FATM followed by a Spiking MLP block, with residual connections applied to output membrane potentials in both blocks. FATM enables multi-scale feature extraction using the proposed spiking frequency representation (Sec~\ref{sec:Frequency Learner}). Lastly, the features from Spiking Encoders undergo Global Average-Pooling (GAP), producing a $D$-dimensional feature, which is then fed into a fully-connected Classification Head (CH) to generate the final prediction Y.

The overall architecture of SWformer is:
\begin{equation}
\begin{array}{ll}
S_0 = \operatorname{\textit{Spk}}(U_0), & S_0 \in \mathbb{R}^{T \times N \times D} \\[0.3em] 
U_l = \operatorname{FATM}(S_{l-1}) + U_{l-1}, &  U_l \in \mathbb{R}^{T \times N \times D},~ l=1,\ldots,M \\ [0.3em] 
S_l= \operatorname{\textit{Spk}}(\operatorname{MLP}(\operatorname{\textit{Spk}}(U_l))+U_l), & S_N \in \mathbb{R}^{T \times N \times D} ~ l=1,\ldots,M \\ [0.3em] 
Y = \operatorname{CH}(\operatorname{GAP}(S_{N}))\\
\end{array}
\end{equation}
where $U_l$ and $S_l$ denotes the membrane potential and spike output of FATM at $l$-th layer, $M$ refers to total number of layers.

\subsection{Frequency-Aware Token Mixer}

We propose the FATM, a novel component designed to facilitate the mixing of tokens across a wide frequency range in SNNs, serving as an alternative to self-attention-based token mixers in Spiking Transformers~\cite{zhou2022spikformer, zhou2023spikingformer, yao2023spike}. As shown in Fig.~\ref{fig:overview}, the FATM operates on all channels concurrently through three parallel branches: (1) Frequency Learner (FL), using spiking wavelet transform for time-frequency domain learning; (2) Spatial Learner (SL), adopting $3 \times 3$ Conv for extracting spatial features; and (3) Channel Mixer (CM), using spiking point-wise convolution that performs cross-channel information fusion. This design is inspired by the effectiveness of wavelet neural operators~\cite{tripura2023wavelet} and the local perception capability of Conv operations.

To enhance computational parallelism and parameter efficiency, we employ the block-diagonal structure in FL by splitting the $d \times d $ weight matrix into k smaller $d/k \times d/k$  matrices (Fig.~\ref{fig:wavelet}). 
Given $k$ weight blocks , the input feature sequence $S_l \in \mathbb{R}^{T \times N \times D}$ is reshaped into $S_l \in \mathbb{R}^{Tk \times N/k \times H \times W}$ and processed by the FATM:
\begin{equation}
\begin{array}{ll}
U_{FL}^{l} = FL(S_l^{'}) & U_{FL}^{l} \in \mathbb{R}^{Tk \times N/k \times H \times W},\ l=1,\ldots,M\\
U_{SL}^{l}= ConvBN(S_l^{'}) & U_{SL}^{l} \in \mathbb{R}^{Tk \times N/k \times H \times W},\ l=1,\ldots,M\\ 
U_{CM}^{l}= ConvBN(S_l^{'}) & U_{CM}^{l}\in \mathbb{R}^{Tk \times N/k \times H \times W} ,\ l=1,\ldots,M\\
U_{FATM}^{l} = U_{FL}^{l} + U_{SL}^{l} + U_{CM}^{l} & U_{FATM}^{l} \in \mathbb{R}^{Tk \times N/k \times H \times W} ,\ l=1,\ldots,M\\
\end{array}
\end{equation}
where $U_{FL}^{l}$, $U_{SL}^{l}$, $U_{CM}^{l}$, represent the membrane potential outputs of the FL, SL, and CM, respectively. After processing, $U_{FATM}^{l}$ is reshaped back to $\in \mathbb{R}^{T \times N \times D}$. The SL and CM use $3 \times 3$ and $1 \times 1$ convolutions, respectively, leveraging CNNs' powerful capabilities to enhance local feature learning. Note that while block-diagonal multiplication is used only in FL, we also reduce channels in SL and CM, thus their parameters decrease linearly with more weight blocks $k$.

\subsection{Frequency Learner}
\label{sec:Frequency Learner}
The FL projects features to a transform domain, weighting and passing specific frequency modes. Specifically, as shown in Fig.~\ref{fig:overview}, Fl converts raw inputs to the time-frequency domain, weighted, and then converted back to the time domain. It incorporates two key designs: a robust spiking frequency representation that links the sparsity of wavelet transform with SNN's binary and sparse signaling property and a modularized weight matrix that enhances parameter efficiency and computational parallelism.

\hspace{0.5cm}\includegraphics[width=0.9\linewidth]{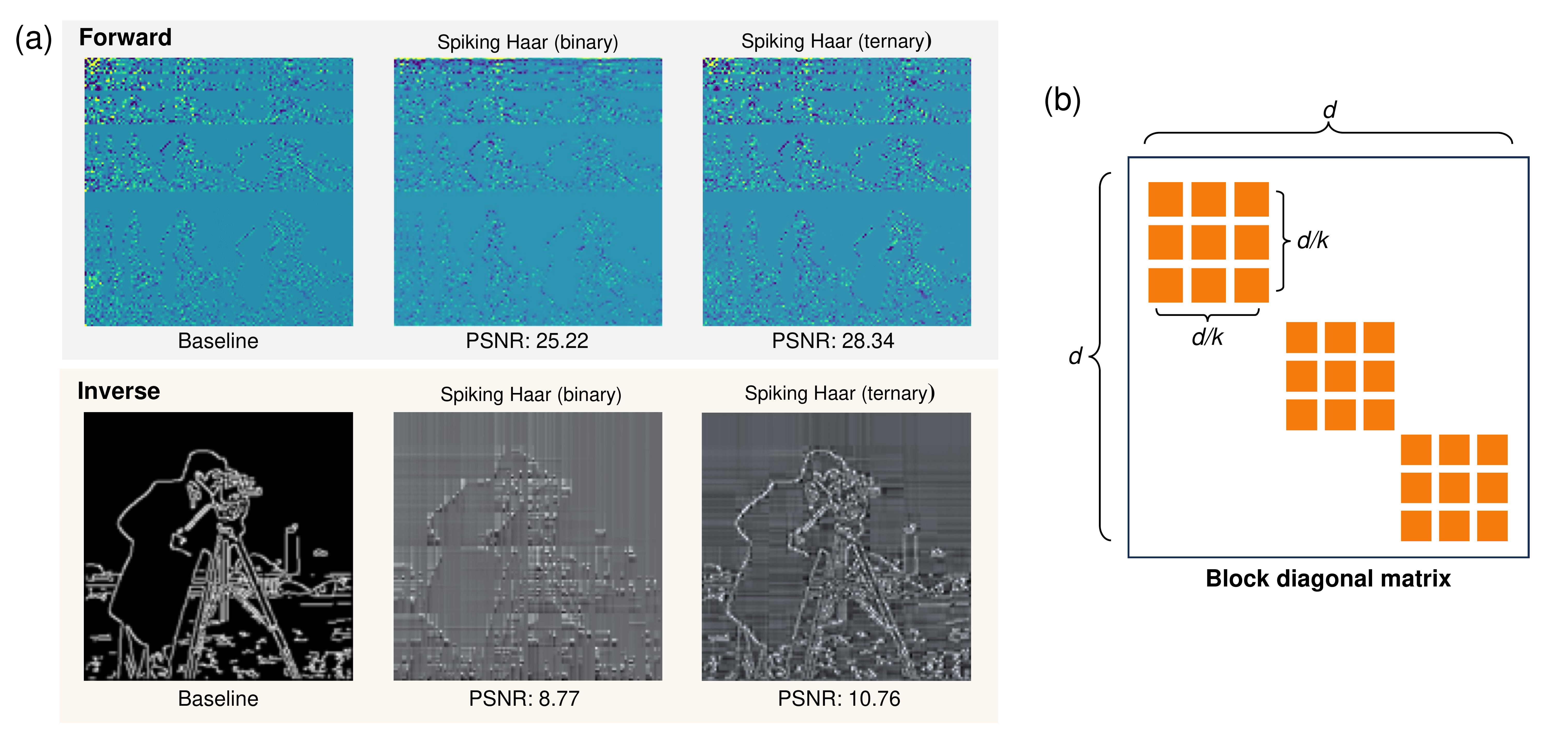}
\caption{ (a) Comparative of the standard Haar transform, binary spiking Haar transform, and ternary spiking Haar transform. Higher Peak Signal-to-Noise Ratio values indicate greater similarity between the images. (b) Schematic of block-diagonal matrix.}
\label{fig:wavelet}

\subsubsection{{{Frequency representation in SNNs}}}~elegantly addresses the challenges of incorporating spike-driven frequency information in SNNs. Our approach is driven by two key insights: (1) the spiking output and reset rules of neuromorphic chips (Fig.~\ref{fig:neuro}) require converting intermediate results to spikes, complicating the direct use of signal transform algorithms with cascaded matrix multiplications; and (2) the amplitude-dependent response of spiking neurons makes algorithms with complex number operations, like the Fourier Transform, resource-intensive, as they need separate neuron banks for real and complex computations. We create the spiking frequency representation using the Haar wavelet transform. This method captures high-frequency details and low-frequency approximations with a sparse representation, while the wavelet transform's decorrelation property enhances signal robustness. The spiking Haar forward and inverse transforms are formulated as:  
\begin{equation}
\begin{aligned}
H_{\text{f}} &= {Spk}( W_{\text{haar}} \cdot {Spk}( I \cdot W_{\text{haar}}^{\top})) \\
I &= {Spk}( W_{\text{haar}}^{\top} \cdot {Spk}( H_{\text{f}} \cdot W_{\text{haar}})) 
\end{aligned}
\end{equation}
\begin{equation}
\begin{aligned}
W_{\text{haar}}(n) &= 
\begin{cases}
[1], & \text{if } n = 1, \\
\frac{1}{\sqrt{2}} 
\begin{bmatrix}
W_{\text{haar}}(n-1) \otimes [1, 1] \\
I_{2^{n-1}} \otimes [1, -1]
\end{bmatrix}, & \text{if } n > 1
\end{cases}
\end{aligned}
\label{eq:8}
\end{equation}
where $I$ and $H_\text{f}$ represent the raw input and the matrix after the Haar forward transform, respectively, and $W_{\text{haar}}$ denotes the transformation matrix. Ideally, the Haar inverse transform recovers $H_\text{f}$ back to $I$ without any error.

Since binary SNNs only generate \{0, 1\} spikes and Eq.~\ref{eq:8} includes negative terms in the Haar transform, causing significant errors, we incorporate negative spike dynamics supported by neuromorphic chips~\cite{davies2018loihi, rathi2023exploring}, expanding spike values to \{-1, 0, 1\}; the ternary neuron model is expressed as:
\begin{align}
U[n] & = V[n-1] + I[n], \\
s[n] & = H_{sym}(U[n-1] - V_{\text{th}}), \\
V[n] & = V_{\text{reset}}s[n] + U[n](1-s[n]),
\end{align}
where H$_{sym}$($\cdot$) refers to a symmetric Heaviside step function, defined as H$_{sym}$(x) = 1 when $x \geq 0$, and H$_{sym}$(x) = -1 when $x < 0$. We adopt the integrate-and-fire neuron~\cite{burkitt2006review}, equivalent to the LIF neuron with $\beta = 1$, for accurate signal transformation. As shown in Fig.~\ref{fig:wavelet}, incorporating negative spike dynamics significantly enhances the quality of the spiking frequency representation. The spiking frequency representation, or spiking wavelet transform, includes forward and inverse transform processes, with negative spike dynamics used only in this part.

 

\subsubsection{{Modularized Weight Matrix}} enables interpretability, computational parallelization, and parameter efficiency, which can be interpreted as batch matrix multiplication~\cite{dao2022monarch}. As illustrated in Fig.~\ref{fig:wavelet}(b), the block-diagonal multiplication approach divides the $d \times d$ weight matrix into $k$ smaller weight blocks, each having a size of $d/k \times d/k$. This technique effectively reduces the parameter count from $O(d^2)$ to $O(d^2/k)$, with better parallelism. With this structure in place, the FL independently processes each weight splitting block as follows: 
\begin{equation}
   {\tilde{y}^{\ell}}_{m, n} = {{W}^{\ell}}_{m, n}{{x}^{\ell}}_{m, n}, ~~\ell=1,..., k, (m,n) \in H \times W
\end{equation}
where $m, n$ refers to the spatial coordinates of a token within the input tensor, and $\ell$ denotes the corresponding block id. Each block can be understood as a head in a multi-head self-attention mechanism, projecting the data into a specific subspace. Choosing an appropriate number of blocks is crucial for obtaining a high-dimensional representation, enabling efficient feature extraction in the frequency spectrum.


\subsection{Membrane Shortcut}

Residual learning and shortcuts are crucial for training deep SNNs~\cite{he2016identity, zheng2021going, fang2021deep, hu2024advancing}. These techniques aim to implement identity mapping to prevent degradation in deep networks while maintaining spike-driven computing for hardware compatibility and energy efficiency.
As shown in Fig.~\ref{fig:shortcut}, three mainstream shortcut techniques are commonly used in SNNs. The Vanilla Shortcut scheme~\cite{zheng2021going} borrows the shortcut scheme from ANNs~\cite{he2016deep}, connecting membrane potential and spike, but it cannot impose identity mapping~\cite{he2016identity}. The Spiking Element-Wise Shortcut~\cite{fang2021deep} connects spikes across layers but operates in an "integer-driven" rather than "spike-driven" manner due to integral output spikes. The Membrane Shortcut~\cite{hu2024advancing} combines identity mapping with spike-driven computation by connecting membrane potentials between layers, optimizing membrane potential distribution. This method, used in recent Spiking Transformers~\cite{yao2023spike}, is also adopted in our SWformer for its biological plausibility and high performance~\cite{yao2023spike, hu2024advancing}.

\hspace{3.0cm}\includegraphics[width=0.52\linewidth]{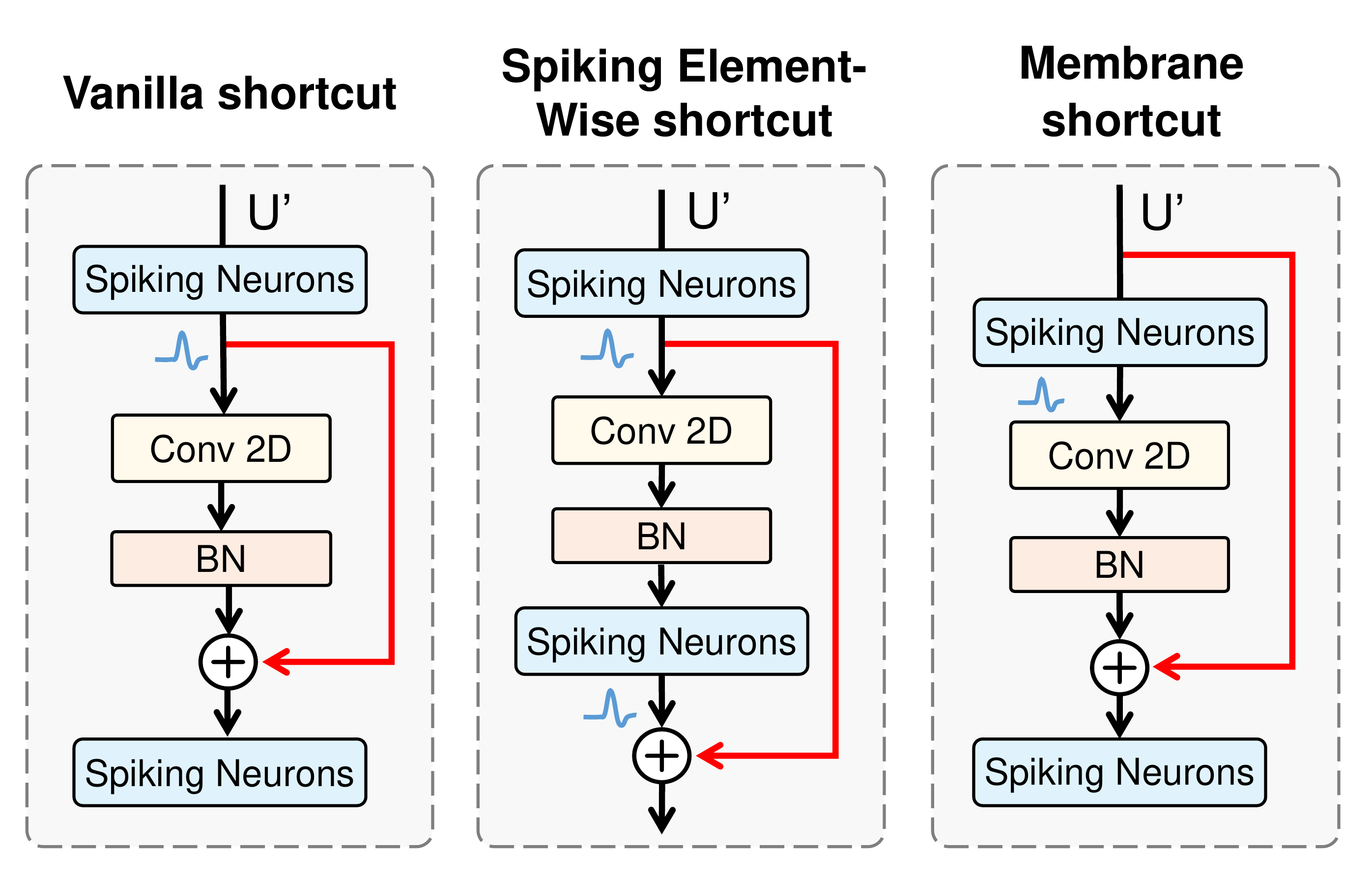}
\caption{{Mainstream shortcut schemes in SNNs.}}
\label{fig:shortcut}

\section{Experiment}

\subsection{Experiment Setup}

SNNs transmit spatio-temporal information, which are naturally suitable for handling temporal tasks. For static image classification, it is common practice to repeatedly input the same image at each timestep. While increasing simulation timesteps can improve accuracy, it also increases training time, hardware requirements, and inference energy consumption. Neuromorphic datasets with inherent spatio-temporal dynamics can fully exploit the energy-efficient advantages of SNNs.

\subsubsection{Dataset} We evaluate our approach on a range of datasets, including static datasets like CIFAR-10~\cite{lecunGradientbasedLearningApplied1998a}, CIFAR-100~\cite{krizhevskyLearningMultipleLayers2009a}, and ImageNet~\cite{dengImagenetLargescaleHierarchical2009}, as well as neuromorphic datasets such as CIFAR10-DVS~\cite{liCIFAR10DVSEventStreamDataset2017}, N-Caltech101~\cite{orchardConvertingStaticImage2015b}, N-Cars~\cite{sironiHATSHistogramsAveraged2018a}, ActionRecognition~\cite{miaoNeuromorphicVisionDatasets2019}, ASL-DVS~\cite{biGraphbasedObjectClassification2019}, and NavGesture~\cite{maro2020event} datasets. Details on training settings and energy consumption evaluation can be found in Supplementary Material.

\subsection{Performance on Static Datasets}
\subsubsection{ImageNet} 
 SWformer, our proposed model, outperforms the vanilla Spiking Transformer (SpikFormer), Spike-driven Transformer (SD Transformer), and other ResNet-based SNNs on ImageNet in terms of accuracy and efficiency. Experiments with different embedding dimensions, transformer blocks, and spiking wavelet settings demonstrate the importance of precise frequency information as shown in Table~\ref{table:imagenet}. SWformer (Block=2) with the Transformer-6-512 setting achieves 74.98\% accuracy, 2.52\% higher than SpikFormer, and 0.87\% higher than SD Transformer. Increasing the number of weight splitting blocks further improves parameter efficiency and power consumption without compromising accuracy. Specifically, SWformer 8-512-block4-V$_{th}$-1 reaches 75.29\% accuracy with 23.14M parameters, surpassing MS-ResNet-34~\cite{hu2021advancing} and SEW-ResNet-34~\cite{fangDeepResidualLearning2021}, which achieve 67.04\% and 69.15\% accuracy, respectively, with approximately 21.8M parameters (22.03\% less than SpikFormer). Note that the baseline using standard wavelet transform, shown in gray, generally achieves better performance, highlighting the importance of accurate frequency representation in SNNs, despite contradicting the spike-driven computing paradigm. To harness the energy efficiency of neuromorphic computing, exploring hardware designed for better frequency domain processing is essential. Recently, the Resonate-and-Fire neuron is supported by Loihi 2, which can compute the Short Time Fourier Transform, another type of time-frequency transform on-chip~\cite{frady2022efficient}.

\begin{table*}[!t]
\begin{center}
\caption{Performance comparison between the proposed model and the SOTA models on the ImageNet dataset. Models denoted with an asterisk (*) use an input resolution of 256$\times$256, which is essential for the Haar transform to achieve optimal performance and consistent application throughout the entire input. All models were trained for 310 epochs with identical initial settings for a fair comparison. {$^{\dagger}$}: B denotes the number of weight splitting blocks. \textbf{$-$}: standard wavelet transform (non-spiking).}
\label{table:imagenet}
\resizebox{\textwidth}{!}{%
\setlength\tabcolsep{6pt} 
\begin{tabular}{ccccccc}
\toprule[1pt]
\multicolumn{2}{c}{\textbf{Methods}} & \textbf{Architecture} & \textbf{\# Param (M)} & \textbf{Power (mJ)} & \textbf{Time Steps} & \textbf{Accuracy (\%)} \\
\midrule
\multicolumn{2}{c}{Hybrid training~\cite{rathi2020enabling}\textsuperscript{ICLR}} & ResNet-34 & 21.79 & - & 250 & 61.48 \\ \cmidrule{3-7}
\multicolumn{2}{c}{\multirow{4}{*}{SEW ResNet~\cite{fang2021deep}\textsuperscript{NeurIPS}}} & SEW-ResNet-34 & 21.79 & 4.04 & 4 & 67.04 \\
& & SEW-ResNet-50 & 25.56 & 4.89 & 4 & 67.78 \\
& & SEW-ResNet-101 & 44.55 & 8.91 & 4 & 68.76 \\
& & SEW-ResNet-152 & 60.19 & 12.89 & 4 & 69.26 \\ \cmidrule{3-7}
\multicolumn{2}{c}{TET~\cite{deng2022temporal}\textsuperscript{ICLR}} & SEW-ResNet-34 & 21.79 & - & 4 & 68.00 \\ \cmidrule{3-7}
\multicolumn{2}{c}{\multirow{2}{*}{MS ResNet~\cite{hu2021advancing}\textsuperscript{TNNLS}}} & MS-ResNet-18 & 11.69 & 4.29 & 4 & 63.10 \\
& & MS-ResNet-34 & 21.80 & 5.11 & 4 & 69.42 \\ \cmidrule{3-7}
\multicolumn{2}{c}{Spiking ResNet~\cite{hu2021spiking}\textsuperscript{TNNLS}} & ResNet-50 & 25.56 & 70.93 & 350 & 72.75 \\ \cmidrule{3-7}
\multicolumn{2}{c}{tdBN~\cite{zhengGoingDeeperDirectlytrained2021}\textsuperscript{AAAI}} & Spiking-ResNet-34 & 21.79 & 6.39 & 6 & 63.72 \\
\midrule
\multicolumn{2}{c}{ANN Transformer*} & Transformer-6-512 & 23.37 & 40.72 & - & 80.54 \\
\midrule
\multicolumn{2}{c}{\multirow{3}{*}{SpikFormer~\cite{zhou2022spikformer}\textsuperscript{ICLR}}} & Transformer-8-384 & 16.81 & 7.73 & 4 & 70.24 \\
& & Transformer-6-512 & 23.37 & 9.41 & 4 & 72.46 \\
& & Transformer-8-512 & 29.68 & 11.57 & 4 & 73.38 \\
\midrule
\multicolumn{2}{c}{\multirow{3}{*}{SD Transformer~\cite{yao2023spike}\textsuperscript{NeurIPS}}} & Transformer-8-384 & 16.81 & 3.39 & 4 & 72.28 \\
& & Transformer-6-512 & 23.37 & 3.56 & 4 & 74.11 \\
& & Transformer-8-512 & 29.68 & 4.50 & 4 & 74.57 \\
\midrule
\multirow{6}{*}{\textbf{SWfomer* (B{$^{\dagger}$}=2)}} & \cellcolor{mygray}\multirow{2}{*}{\textbf{$-$}} & \cellcolor{mygray}Transformer-6-512 & \cellcolor{mygray}21.8 & \cellcolor{mygray}4.00 & \cellcolor{mygray}4 & \cellcolor{mygray}75.09 \\
& \cellcolor{mygray}& \cellcolor{mygray}Transformer-8-512 & \cellcolor{mygray}27.6 & \cellcolor{mygray}4.31 & \cellcolor{mygray}4 & \cellcolor{mygray}75.26 \\ \cmidrule{2-7}
& \multirow{2}{*}{V$_{th}$=0.5} & Transformer-6-512 & 21.8 & 3.58 & 4 & 74.98 \\
& & Transformer-8-512 & 27.6 & 4.89 & 4 & 75.18 \\ \cmidrule{2-7}
& \multirow{2}{*}{V$_{th}$=1} & Transformer-6-512 & 21.8 & 3.87 & 4 & 74.84 \\
& & Transformer-8-512 & 27.6 & 5.08 & 4 & 75.43 \\ \midrule
\multirow{6}{*}{\textbf{SWfomer* (B{$^{\dagger}$}=4)}} & \cellcolor{mygray}\multirow{2}{*}{\textbf{$-$}} & \cellcolor{mygray}Transformer-6-512 & \cellcolor{mygray}18.46 & \cellcolor{mygray}3.51 & \cellcolor{mygray}4 & \cellcolor{mygray}74.86 \\
& \cellcolor{mygray}& \cellcolor{mygray}Transformer-8-512 & \cellcolor{mygray}23.14 & \cellcolor{mygray}4.67 & \cellcolor{mygray}4 & \cellcolor{mygray}75.33 \\ \cmidrule{2-7}
& \multirow{2}{*}{V$_{th}$=0.5} & Transformer-6-512 & 18.46 & 3.91 & 4 & 74.62 \\
& & Transformer-8-512 & 23.14 & 4.98 & 4 & 75.08 \\\cmidrule{2-7}
& \multirow{2}{*}{V$_{th}$=1} & Transformer-6-512 & 18.46 & 3.75 & 4 & 74.69 \\
& & Transformer-8-512 & 23.14 & 4.87 & 4 & 75.29 \\
\bottomrule[1pt]
\end{tabular}
}
\end{center}
\end{table*}

\subsubsection{CIFAR10/ CIFAR100 } 
Table~\ref{table:cifar} presents a comprehensive comparison of the SWformer model with current state-of-the-art (SOTA) SNN models on the CIFAR-10/100. SWformer outperforms all other models in terms of top-1 accuracy on both datasets with fewer parameters and time steps. In specific, compared to ResNet-based SNN models like tdBN~\cite{zhengGoingDeeperDirectlytrained2021}, our SWformer model outperforms it by {2.0\%} on CIFAR10 and {5.4\%} on CIFAR100, with only {59.5\%} of the parameters. Additionally, the SWformer model also surpasses all the Transformer-based SNNs in accuracy and parameter efficiency. The superior performance of SWformer can be attributed to its unique designs. 

\begin{table*}[!t]
\begin{center}
\caption{Performance comparison on CIFAR10/CIFAR100 and CIFAR10-DVS.}
\label{table:cifar}
\resizebox{\textwidth}{!}{
\setlength\tabcolsep{6pt} 
\begin{tabular}{@{}cccccccccc@{}}
\toprule
\multirow{2}{*}{\textbf{Method}} & \multicolumn{3}{c}{\textbf{CIFAR10}} & \multicolumn{3}{c}{\textbf{CIFAR100}} & \multicolumn{3}{c}{\textbf{CIFAR10-DVS}}\\

\cmidrule(r){2-4} \cmidrule(l){5-7} \cmidrule(l){8-10}
& \textbf{\# Param (M)} & \textbf{T} & \textbf{Acc. (\%)} & \textbf{\# Param (M)}& \textbf{T} & \textbf{Acc. (\%)} & \textbf{\# Param (M)} & \textbf{T} & \textbf{Acc. (\%)} \\
\midrule
TET~\cite{deng2022temporal} \textsuperscript{ICLR}& 12.63 & 6 & 94.50 & 12.63 &6 &74.72 & 9.27 & 10 & 83.32 \\ \cline{2-10}
tdBN~\cite{zhengGoingDeeperDirectlytrained2021} \textsuperscript{AAAI}& 12.63& 4 & 92.92 & 12.63& 4 & 70.86 & 12.63 & 10 & 67.8 \\ \cline{2-10}
TEBN~\cite{duan2022temporal} \textsuperscript{NeurIPS}& 12.63 & 6 & 94.71 & 12.63& 6 & 76.41 & - & 10 & 75.10 \\ \cline{2-10}
Real Spike~\cite{guo2022real} \textsuperscript{ECCV}& 12.63 & 6 & 95.78 & 39.9 & 10 & 71.24 & 12.63 & 10 & 72.85 \\ \cline{2-10}
DSR~\cite{mengTrainingHighPerformanceLowLatency2022} \textsuperscript{CVPR}& 11.2  & 20 & 95.4 & 11.2 & 20 & 78.5 & 9.48 & 10 & 77.51\\ \cline{1-10}
\addlinespace
\multirow{2}{*}{SpikFormer~\cite{zhou2022spikformer} \textsuperscript{ICLR}} & 9.32 & 4 & 95.51 & 9.32 & 4 & 78.21 & 2.59 & 10 & 78.9 \\ 
& 9.32 & 6 & 95.34 & 9.32 & 4 & 78.61 & 2.59 & 16 & 80.9 \\  \cline{1-10}
\addlinespace
\multirow{1}{*}{SD Transformer ~\cite{yao2023spike} \textsuperscript{NeurIPS}} & 9.32 & 4 & 95.6 & 9.32 & 4 & 78.4 & 2.59 & 10 & 78.9 \\ \cline{1-10}
\addlinespace
\multirow{2}{*}{\textbf{SWformer}} & \textbf{7.51} & \textbf{4} & \textbf{96.1} & \textbf{7.51} & \textbf{4} & \textbf{79.3} & \textbf{2.05} & \textbf{10} & \textbf{82.9}\\

& \textbf{7.51} & \textbf{6} & \textbf{96.3} & \textbf{7.51} & \textbf{6} & \textbf{79.6} & \textbf{2.05} & \textbf{16} & \textbf{83.9}\\
\bottomrule
\end{tabular}
}
\end{center}

\end{table*}

\begin{table}[!t]
    \centering
    \begin{minipage}{0.5\linewidth} 
        \centering
        \caption{Performance comparison of SWformer vs. SOTA on neuromorphic datasets.}
        \label{table:event}
        \vskip -0.1in
        \setlength{\tabcolsep}{15pt}
        \resizebox{0.9\linewidth}{!}{
        \begin{tabular}{cccc}
            \toprule
            \textbf{Datasets} & \textbf{Methods} & \textbf{T} & \textbf{Acc. (\%)} \\
            \cmidrule{1-4}
            \multirow{6}{*}{N-CALTECH101} 
                                          & TIM~\cite{shen2024tim} & 10 & 79.00 \\
                                          & TT-SNN~\cite{lee2024tt} & 6 & 77.00 \\
                                          & NDA~\cite{liNeuromorphicDataAugmentation2022} & 10 & 83.70 \\
                                          \cmidrule{2-4}
                                          & \cellcolor[gray]{0.9} \textbf{SWformer} & \cellcolor[gray]{0.9}10 & \cellcolor[gray]{0.9}\textbf{88.45} \\ \cmidrule{1-4}
            \addlinespace
            \multirow{3}{*}{N-CARS} & CarSNN~\cite{vialeCarsnnEfficientSpiking2021} & 10 & 86.00 \\
                                    & NDA~\cite{liNeuromorphicDataAugmentation2022} & 10 & 91.90 \\
                                    \cmidrule{2-4}
                                    & \cellcolor[gray]{0.9} \textbf{SWformer} & \cellcolor[gray]{0.9}10 & \cellcolor[gray]{0.9}96.32\\\cmidrule{1-4}
                                    
\multirow{3}{*}{Action Recognition}      & STCA~\cite{guSTCASpatiotemporalCredit2019}        & 10                           & 71.20                \\
                              & Mb-SNN ~\cite{liuEventbasedActionRecognition2021a}          & 10                               & 78.10             \\
                              \cmidrule{2-4}
                              &\cellcolor[gray]{0.9}\textbf{SWformer}       & \cellcolor[gray]{0.9}10                                               & \cellcolor[gray]{0.9}\textbf{88.88}    \\ 
                              \cmidrule{1-4} 
\multirow{2}{*}{ASL-DVS}      & Meta-SNN~\cite{stewartMetalearningSpikingNeural2022}        & 100                           &   96.04             \\
                              \cmidrule{2-4}
                              &\cellcolor[gray]{0.9}\textbf{SWformer}       & \cellcolor[gray]{0.9}10                                               & \cellcolor[gray]{0.9}\textbf{99.88}    \\ 
                              \cmidrule{1-4} 
\multirow{2}{*}{NavGesture}      & KNN~\cite{maro2020event}        & -                           &   95.90            \\
                              \cmidrule{2-4}
                              &\cellcolor[gray]{0.9}\textbf{SWformer}       & \cellcolor[gray]{0.9}10                                               & \cellcolor[gray]{0.9}\textbf{98.49}    \\ 
            \bottomrule  
        \end{tabular}
        }
        
    \end{minipage}
    \hfill
    \begin{minipage}{0.46\linewidth} 
        \centering
        \caption{Ablation study on FATM}
        \label{tab:ablation}
        \vskip -0.15in
        \resizebox{\linewidth}{!}{
       
        \begin{tabular}{llcc}
            \toprule
            \textbf{Datasets} & \textbf{Models} & \textbf{T} & \textbf{Acc. (\%)} \\
            \midrule
\multirow{5}{*}{CIFAR100} 
                             & SWformer-4-384\textsubscript{NoHaar}  & 4 &  78.89\\
                             & {SWformer-4-384\textsubscript{MaskDC}}  & 4 &   79.34\\
                             & SWformer-4-384\textsubscript{NoNeg} & 4 & 78.64 \\
                             & {SWformer-4-384\textsubscript{NoInv}}  & 4 &   78.79\\
                             & {SWformer-4-384\textsubscript{base}}  & 4 &   79.31\\
                              
\midrule
\multirow{4}{*}{CIFAR10-DVS}    
                            & SWformer-2-256\textsubscript{NoHaar}  & 16 & 80.9 \\
                             & SWformer-2-256\textsubscript{MaskDC}  & 16 & 84.0 \\
                            & SWformer-2-256\textsubscript{NoPool}     & 16 & 81.8 \\
                             & {SWformer-2-256\textsubscript{base}}   & 16 & 83.9 \\
            \bottomrule
        \end{tabular}
        }
        
    \end{minipage}
\end{table} 

\begin{figure*}[t!]
 
   \hspace{-0.2cm}\vspace{0cm}\includegraphics[width=1.0\linewidth]{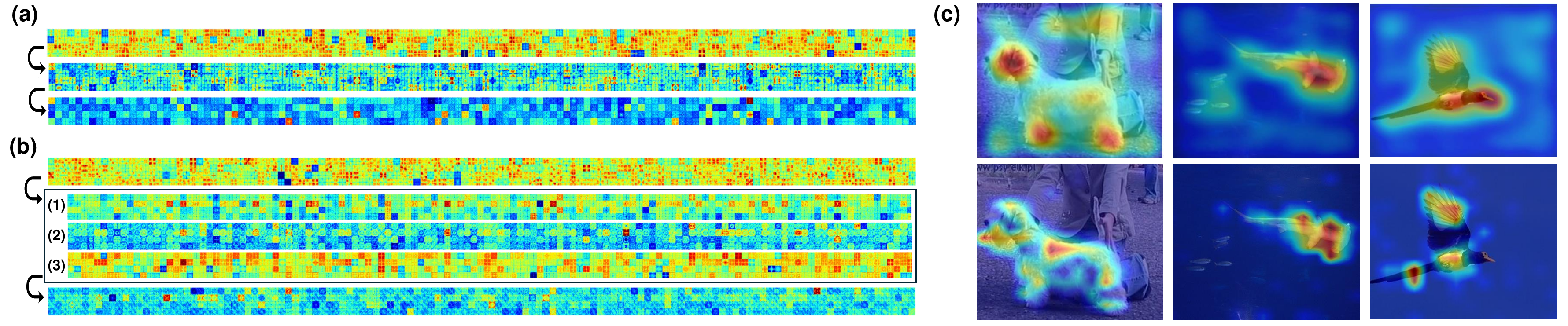}
 \caption{{ (a-b) Fourier analysis of feature maps on ImageNet for the output of SPS (top), the first token-mixer (middle), and the last layer (bottom): (a) global Spiking Self-Attention (gSSA), (b) FATM: (1-3) CM, SL, and FL. FATM is more effective at capturing frequency information than gSSA, which ensures feature perception across a wide frequency range.
  (c) Grad-CAM~\cite{selvaraju2017grad} activation map visualization of the gSSA block in Spiking Transformer\cite{yao2023spike} (top), and FATM in SWformer (bottom).}}
	\label{fig:analysis}
\end{figure*}


\subsection{Performance on Neuromorphic Datasets}

As shown in Table~\ref{table:cifar} and Table~\ref{table:event}, the proposed SWformer outperforms SOTA SNN models on a variety of neuromorphic datasets, including CIFAR10-DVS~\cite{liCIFAR10DVSEventStreamDataset2017}, N-Caltech101~\cite{orchardConvertingStaticImage2015b}, and N-Cars~\cite{sironiHATSHistogramsAveraged2018a}, which are derived from static datasets and converted into neuromorphic data using event-based cameras. For CIFAR10-DVS~\cite{liCIFAR10DVSEventStreamDataset2017}, we revise the FATM shortcut: inputs are first processed by FL, which acts as a filter, and then by SL and CM. To further enhance high-frequency information in these datasets, a 1D max-pooling layer is placed at the beginning of the FATM module. SWformer achieves impressive accuracy on all neuromorphic tasks. These results surpass previous SOTA models by significant margins, with SWformer outperforming NDA~\cite{liNeuromorphicDataAugmentation2022} by 4.75\% and 4.42\% on N-CALTECH101 and N-CARS, Mb-SNN~\cite{liuEventbasedActionRecognition2021a} by 10.78\% on Action Recognition, Meta-SNN~\cite{stewartMetalearningSpikingNeural2022} by 3.84\% while using 10 times fewer timesteps on ASL-DVS, and KNN~\cite{maro2020event} by 2.59\% on NavGesture.

 

\subsection{Method Analysis}
\label{sec:4.4}
\subsubsection{Visualization}
Existing Spiking Transformers~\cite{zhou2022spikformer,zhou2023spikingformer, yao2023spike} use global operations for exchanging information among non-overlapping patch tokens, while spiking neurons transmit pixel-level brightness changes, enriching images with local information, i.e., high-frequency components. This is supported by Fig.~\ref{fig:analysis}(a-b), where data processed by the SPS module, the initial embedding operation before subsequent Transformer blocks, contains rich high-frequency information. While the Spiking Transformer's gSSA primarily focuses on low frequencies~(Fig.~\ref{fig:analysis}(a)), SWformer's FATM effectively captures specific frequency information on each channel, enabling comprehensive feature learning in the frequency spectrum and maintaining high-frequency information transmission even in deeper layers (Fig.~\ref{fig:analysis}(b)) This enhanced frequency learning capability facilitates more accurate and complete feature extraction, as shown in Fig.~\ref{fig:analysis}(c), leading to improved recognition capability.

\subsubsection{Number of Weight Splitting Blocks}
An appropriate number of weight splitting blocks in SWformer can lead to a more efficient architecture that strikes a better balance between performance and resource utilization, correlating with the frequency mixing range of the transformed signals. Table~\ref{table:imagenet} demonstrates that increasing the weight splitting blocks from 2 to 4 in SWformer reduces both parameters and power consumption while maintaining high accuracy, as illustrated by the computational scheme in Fig.~\ref{fig:wavelet}(b).
Specifically, SWformer (Block=4) has about 16\% fewer parameters than SWformer (Block=2) for both Transformer-6-512 and Transformer-8-512 architectures. Despite the reduced resources, SWformer (Block=2) and SWformer (Block=4) achieves accuracy improvements of 2.52\% and 2.16\% for Transformer-6-512 and 2.05\% to 1.91\% for Transformer-8-512 compared to SpikFormer. These findings highlight the importance of the number of weight splitting blocks in SWformer to create an efficient architecture without compromising performance.

\subsubsection{Firing Threshold of Spiking Frequency Representation}

Spiking neurons act as temporal pruning for inputs, leading to better energy efficiency. As shown in Fig.~\ref{fig:wavelet} and Table~\ref{table:imagenet}, the inherent sparsity and robustness of the wavelet transform, combined with  negative spike dynamics, enable a significant reduction in power consumption without compromising accuracy. Increasing the V$_{th}$ in SWformer (Block=4) reduces power consumption by 58.4\% to 60.1\%. This effect depends on the specific V$_{th}$ value and SWformer architecture setting. Besides, the results using standard wavelet transform, without inserted spiking layers, represented by the gray background data, generally achieves better performance, emphasizing the importance of precise frequency representation in SNNs, albeit contradicting the spike-driven computing paradigm. Therefore, our spiking frequency representation, which combines time-frequency transform with SNNs, is crucial for the entire model design, enabling robust signal projection in just 4 timesteps, as shown in Table~\ref{table:imagenet} and Fig.~\ref{fig:wavelet}.

\subsubsection{Ablation Study of Frequency-Aware Token Mixer}
To better understand the advantages of the FATM, we performed ablation studies. Removing spiking wavelet transforms significantly decreased performance on CIFAR100 and CIFAR10-DVS, highlighting the critical role of frequency feature learning. Interestingly, masking the DC components actually improves performance, highlighting the significance of high-frequency information. Furthermore, on the CIFAR10-DVS dataset, removing the 1D max-pooling operation leads to a performance drop from 83.9\% to 81.8\%. We also assessed the effectiveness of the spiking frequency representation. Evaluating CIFAR100 without negative spike dynamics and spiking Haar inverse transform results in performance reductions from 79.31\% to 78.64\% and 78.79\%, respectively. These findings underscore the vital importance of accurate frequency representation in FATM for optimal performance and emphasize the crucial role of high-frequency components SNNs.



\section{Conclusion}
In this work, we develop the Spiking Wavelet Transformer (SWformer), a powerful alternative to self-attention-based token mixers, with promising performance and parameter efficiency. The core innovation of SWformer is its Frequency-Aware Token Mixer~(FATM), which combines spatial learner~(SL), frequency learner~(FL), and channel mixing~(CM) branches. This unique design enables SWformer to emphasize high-frequency components and enhance the perception capability of Spiking Transformers in the frequency spectrum. Furthermore, we introduce a novel spiking frequency representation that facilitates robust, multiplication-free, and event-driven signal transform. Extensive experiments show that SWformer surpasses representative SNNs on both static and neuromorphic datasets, underscoring the crucial role of frequency learning in spiking neural networks. We believe this study offers the community valuable insights for designing efficient and effective SNN architectures.


\section*{Acknowledgements}
This work is supported by the Guangzhou-HKUST(GZ) Joint Funding Program (Grant No. 2023A03J0682) and partially supported by a collaborative project with Brain Mind Innovation, inc. Special thanks to Mr. Yijian He.

%
%
\bibliographystyle{splncs04}
\bibliography{main}
\end{document}